\documentclass[runningheads]{llncs}
\usepackage{graphicx}
\usepackage{algorithm}
\usepackage{algorithmicx, algpseudocode}
\usepackage{amsmath}
\usepackage{amssymb}
\usepackage{thmtools}
\usepackage{mathtools}
\usepackage{subfigure}
\usepackage{multirow}
\usepackage{wrapfig}
\usepackage[misc,geometry]{ifsym}
\usepackage{booktabs}

\begin{document}
\title{Expanding Semantic Knowledge for Zero-shot Graph Embedding}
%
%
\author{Zheng Wang\inst{1,2} \and
Ruihang Shao\inst{2} \and
Changping Wang\inst{3} \and
Changjun Hu\inst{2} \and
Chaokun Wang\inst{4} \and
Zhiguo Gong\inst{1}}
\authorrunning{Wang et al.}
%
\institute{
State Key Laboratory of Internet of Things for Smart City and Department of Computer and Information Science, University of Macau, Macao, China \and
Department of Computer Science and Technology, University of Science and Technology Beijing, Beijing, China \and
Kwai Inc. \and
School of Software, Tsinghua University, Beijing, China\\
\email{wangzheng@ustb.edu.cn, fstzgg@um.edu.mo}
}
\maketitle              
\begin{abstract}
Zero-shot graph embedding is a major challenge for supervised graph learning.
Although a recent method RECT has shown promising performance, its working mechanisms are not clear and still needs lots of training data.
In this paper, we give deep insights into RECT, and address its fundamental limits.
We show that its core part is a GNN prototypical model in which a class prototype is described by its mean feature vector.
As such, RECT maps nodes from the raw-input feature space into an intermediate-level semantic space that connects the raw-input features to both seen and unseen classes.
This mechanism makes RECT work well on both seen and unseen classes, which however also reduces the discrimination.
To realize its full potentials, we propose two label expansion strategies.
Specifically, besides expanding the labeled node set of seen classes, we can also expand that of unseen classes.
Experiments on real-world datasets validate the superiority of our methods.
\end{abstract}

\section{Introduction}
Graph embedding is becoming a major trend among various graph processing methods~\cite{wang2018deepdirect}~\cite{xiao2014user}.
Most recently, there has been an increasing interest in supervised graph embedding~\cite{kipf2017semi}.
However, little work has considered the \emph{zero-shot graph embedding (ZGE)} problem where some classes have no labeled data at the training time.
This problem has practical significance, especially in domains where the graph size is typically large and node class labels can take on many values.
Moreover, general supervised methods would deliver very unsatisfying results in this setting.

To fix this problem, RSDNE~\cite{wang2018rsdne} relaxes the constraints of intra-class similarity and inter-class dissimilarity, so as to avoid the negative influence of missing the labeled data from unseen classes.
However, this method cannot model the high non-linear information or the rich information of a graph.
A recently proposed graph neural network (GNN)~\cite{scarselli2008graph} method named RECT~\cite{wang2020RECT} overcomes these limits, having shown favorable performance.
Nevertheless, its working mechanisms are still not clear, significantly hindering its practicality.

In this paper, we demystify the RECT for ZGE.
In particular, we show that its core part (named RECT-L) can be thought as a GNN prototypical model which learns a nearest class mean (NCM) classifier~\cite{webb2003statistical}.
This explains why RECT works on seen classes.
On the other hand, the learned prototypical model maps nodes from the raw-input feature space into a ``semantic'' space where a class is described by its mean feature vector.
This enables transferring knowledge from seen classes to unseen classes, which is the fundamental reason why RECT works well on the nodes coming from unseen classes.
However, it also leads to the ineffectiveness of RECT, as semantic knowledge contains much less discriminative information than the original binary labels.

To overcome this limit and realize the full potentials of RECT, we design two label expansion strategies.
The first is to expand the labeled node set of seen classes, which will make RECT ``see'' more labels.
This overcomes the localized nature of the used GNN model~\cite{li2018deeper}.
The other one is to jointly expand the labeled node sets of both seen and unseen classes.
This improves the diversity of labels, which would yield more robust embedding results.
Combining these two strategies can substantially improve the performance of RECT, especially when the labeled data is very limited.
In addition, we further provide some theoretical analysis for the proposed expansion strategies.
Finally, we conduct extensive experiments to demonstrate the effectiveness of our methods.

\section{Why RECT Work}
\subsection{Problem Definition}
The problem of zero-shot graph embedding (ZGE) in this paper follows~\cite{wang2018rsdne}.
A graph generally consists of a set of nodes that are possibly connected by edges.
We are given a labeled training node set $\mathcal{L}$ whose label set is $\mathcal{C}^{s}$ (i.e., the seen class set).
The rest are testing nodes some of which come from an unseen class set $\mathcal{C}^{u}$, i.e., $\mathcal{C}^{s} \cap \mathcal{C}^{u} = \emptyset$.
By using the labeled nodes only from $\mathcal{C}^{s}$ where no labeled nodes of $\mathcal{C}^{u}$ is available, we aim to learn low-dimensional node representation vectors, such that the nodes with similar neighbors, features, or labels are close to each other in the learned embedding space.

\subsection{Preliminaries: RECT}\label{preli_section}
RECT contains two sub-parts: RECT-N and RECT-L, both of which utilize GNN~\cite{scarselli2008graph} layers for embedding learning.
The first part RECT-N is unsupervised, aiming to preserve the original graph structure.
The other and most noteworthy part is the supervised method RECT-L.
Inspired by the success of ZSL, RECT-L learns with the class-semantic descriptions of seen classes, i.e., semantic knowledge is introduced for transferring supervised knowledge from seen to unseen classes.
Unlike traditional ZSL methods whose semantic knowledge is human annotated or provided by some third-party resources (like the word2vec tools~\cite{mikolov2013efficient}), RECT-L obtains this knowledge in a practical domain-dependent manner with a ``readout'' function.
Specifically, for each seen class $c$, it uses the mean feature of all corresponding nodes in this class as its class-semantic description vector $\hat{y}_{c}$: $\hat{y}_{c} = \mathrm{MEAN}(\{ x_{i}|\forall_{i}\ \mathcal{C}^{s}_{i}=c \})$,
where $x_i$ and $\mathcal{C}^{s}_{i}$ are node $i$'s feature vector and seen class label, respectively.
Finally, RECT-L minimizes the difference between the predicted and the actual class-semantic description vectors:
\begin{equation}
\label{eq_semantic_loss}
\mathcal{J} = \sum_{i\in \mathcal{L}} \ell (\hat{y}'_{C_{i}^{s}}, \hat{y}_{C_{i}^{s}})
\end{equation}
where $\hat{y}'_{C_{i}^{s}}$ and $\hat{y}_{C_{i}^{s}}$ stand for the predicted and actual class-semantic vector of node $i$ respectively, and $\ell(\cdot, \cdot)$ is a sample-wise loss function.

\subsection{RECT-L v.s. ZSL Methods}\label{sub_sect_relationship}
Theoretically, a typical ZSL method can be thought of a semantic output code classifier $\mathcal{F}: X^{d} \to Y$, such that $\mathcal{F}$ contains two other functions, $\mathcal{S}$ and $\mathcal{Q}$~\cite{palatucci2009zero}:
\begin{equation}\label{eq_gnn}
\begin{aligned}
&\mathcal{F}\ =\ \mathcal{Q}(\mathcal{S}(\cdot)) \\
&\mathcal{S}:\ X^{d} \to Z^{p} \\
&\mathcal{Q}:\ Z^{p} \to Y
\end{aligned}
\end{equation}
where $\mathcal{S}$ is a semantic mapping function which maps from a $d$-dimensional raw-input space $X^d$ into a $p$-dimensional semantic space $Z^p$; and
$\mathcal{Q}$ is a semantic decoding function which maps the obtained semantic encoding to a class label from a label set $Y$.
The classifier $\mathcal{F}$ is given a knowledge base $\mathcal{K}$ which guides the learning of $\mathcal{S}$ and $\mathcal{Q}$.
Practically, $\mathcal{K}$ is usually simplified as a one-to-one encoding between class labels and semantic space points.
A commonly used encoding is: a class label and its corresponding class-semantic description vector.

In RECT-L, a class (prototype) is described by its mean feature vector, indicating the used semantic space is directly constructed from the $d$-dimensional raw-input features.
As such, the knowledge base $\mathcal{K}$ could only guide the learning of semantic mapping function $\mathcal{S}$ rather than the semantic decoding function $\mathcal{Q}$.
This is because only the one-to-one encoding between seen class labels and semantic space points (i.e., a seen class and its mean feature vector) is known in ZGE problem.
In other words, $\mathcal{K}$ does not contain any knowledge about the relationship between semantic space points and unseen classes, since it is impossible to obtain the mean feature vectors of unseen classes when there exists no labeled nodes from unseen classes.
This is the fundamental difference between RECT-L and ZSL methods.
\begin{remark}[The Difference Between RECT-L and ZSL Methods]\label{remark_relationship}
In the semantic space of ZSL methods, class prototypes are described by human annotation or third-part resources; while in the semantic space of RECT-L, class prototypes are described by their mean feature vectors.
In addition, in RECT-L, the knowledge of relationship between unseen classes and semantic space points is unknown.
\end{remark}

\subsection{The Mechanisms of RECT}
We continue with it's core part RECT-L.
As analysed above, RECT-L adopts GNN layers and finally ends with a semantic loss (i.e., Eq.~\ref{eq_semantic_loss}), where class prototypes are represented by their mean feature vectors.
From the viewpoint of classification theory, this is the NCM classifier loss~\cite{mensink2012metric}.

\begin{remark}[The Reasonability of RECT-L]\label{remark_reason}
As shown above, RECT-L actually learns a prototypical model with the labeled data of seen classes, reflecting its reasonability on seen classes.
On the other hand, as shown in Remark~\ref{remark_relationship}, the learned prototypical model maps the data from the raw-input space into a semantic space, like ZSL methods.
As validated by lots of ZSL methods, this enables the success of transferring supervised knowledge of seen classes to unseen classes, indicating its reasonability on unseen classes.
\end{remark}

\section{How to Improve RECT}

\subsection{The Proposed Method}
We overcome the limit of RECT by designing two label expansion strategies.
The first is to expand the seen class label set.
As directly learning with the binary labels would get unappealing results in ZGE problem~\cite{wang2018rsdne}, we preform label expansion based on the semantic method RECT-L.
This naturally leads to a self-training strategy.
Specifically, we first train a RECT-L model as described in Section~\ref{preli_section}.
Then, we use the learned model to get the predicted class-semantic descriptions of unlabeled nodes.
After that, for each seen class, we can find top $k$ closest unlabeled nodes to its class-semantic description vector in the semantic space, and finally add them to the labeled node set of this class.

The other is to expand both the seen and unseen class label sets.
This would improve the diversity of labels and obtain more robust node embeddings.
Although we know little about unseen classes, we can still find some ``labeled'' data for them.
Our idea is quite simple: exploring the discriminative information of both seen and unseen classes via clustering.
Specifically, we first train a RECT-L model to get the node embeddings.
Then, we apply K-means clustering on the resulted embeddings.
After that, for each cluster (class), we can find top $k$ nearby nodes w.r.t. each class center, and finally use them as the labeled data of this class.
As K-means clustering is performed on all classes, the newly obtained labeled node set is expected to cover all of them.

\subsection{Risk Bounds Analysis}\label{sub_sect_risk}
We apply the related learning theories in domain adaptation~\cite{ben2007analysis} to our method.
Let $D^{train} = \{D^{train}_{original} \cup D^{train}_{expand} \} $ denote the final labeled training node set, where $D^{train}_{original}$ denotes the original labeled node set and $D^{train}_{expand}$ denotes the newly added labeled set via label expansion.
Let $D^{test} = D - D^{train}_{original}$ denote the testing node set, where $D$ is the whole node set.
The distribution of $D^{train}$ is $P_{train}$ and of $D^{test}$ is $P_{test}$.
The true class-semantic description labeling function is $h(x)$ and the learned prediction function is $f(x)$.
We define the prediction error in $D^{train}$ and $D^{test}$ as:
\begin{equation}
\label{eq_prediction_error}
\begin{aligned}
\epsilon_{train}(f) & = \mathbb{E}_{x\sim P_{train}} [| h(x) - f(x) |]\\
\epsilon_{test}(f)  & = \mathbb{E}_{x\sim P_{test}} [| h(x) - f(x) |]
\end{aligned}
\end{equation}

We can consider it as a domain adaptation problem.
Suppose the hypothesis space $\mathcal{H}$ containing $f$ is of VC-dimension $\bar{d}$.
According to Theorem 1 in~\cite{ben2007analysis}, with probability at least $1{-}\delta$, for every $f \in \mathcal{H}$, the expected error $\epsilon_{test}(f)$ is bounded:
\begin{equation}
\label{eq_error_bound}
\begin{aligned}
\epsilon_{test}(f) & \le \hat{\epsilon}_{train}(f) + \sqrt{\frac{4}{l}(\bar{d}\log{\frac{2el}{\bar{d}}} + \log{\frac{4}{\delta}} )} \\
 & + d_{\mathcal{H}}(D^{train}, D^{test}) + \rho
\end{aligned}
\end{equation}
where $\hat{\epsilon}_{train}(f)$ is the empirical error of $f$ in $D^{train}$, $e$ is the base of natural logarithm, $l$ is the labeled node number after label expansion, $\rho = \inf_{h \in \mathcal{H}}[\epsilon_{train}(f)+\epsilon_{test}(f)]$, and $d_{\mathcal{H}}(D^{train}, D^{test})$ is the distribution distance between $D^{train}$ and $D^{test}$.

The first term in Eq.~\ref{eq_error_bound} is explicitly minimized by training with $D^{train}$ in Eq.~\ref{eq_semantic_loss}.
If we have high quality $D^{train}_{expand}$, it is expected that we can learn a model that has a small error on $D^{train}$. On the other hand, the bad $D^{train}_{expand}$, e.g., random labels, may lead to a large empirical error.
For the second term, we can notice that the final labeled node number $l$ (after label expansion) is definitely larger than the original one.
This verifies the reasonability of our label expansion strategy.
The third term reflects the relatedness between training and testing data.
In the best situation where $D^{train}$ and $D^{test}$ have the same conditional distribution given a class, and suppose all instances are i.i.d., the distribution distance will be small.
Besides, introducing more correctly labeled nodes will also reduce this distance~\cite{hastie2009elements}, as we have $D^{train}_{expand} \subseteq D^{test}$.

\section{Experiments}
In this section, we conduct extensive experiments to demonstrate the effectiveness of our methods:
1) Ours$_{\mathit{SL}}$: only expanding the labeled node set of seen classes;
2) Ours$_{\mathit{SUL}}$: expanding the labeled node sets of both seen and unseen classes, when the real class number is given;
3) Ours$_{\mathit{SUL}^*}$: expanding the labeled node sets of both seen and unseen classes, when the real class number is estimated automatically\footnote{The optimal class number is determined by silhouette coefficient~\cite{rousseeuw1990finding}.};
4) Ours$_{\mathit{SL\textit{-}SUL}}$: concatenating the embeddings obtained by Ours$_{\mathit{SL}}$ and Ours$_{\mathit{SUL}}$;
and 5) Ours$_{\mathit{SL\textit{-}SUL}^*}$: concatenating the embeddings obtained by Ours$_{\mathit{SL}}$ and Ours$_{\mathit{SUL}^*}$.

\subsection{Setup}
\begin{table}[!t]
\setlength{\tabcolsep}{10pt} 
\caption{The Statistics of Datasets.}
\centering
\begin{tabular}{lrrrr}
\toprule
\textbf{Dataset}     & \textbf{Nodes}     & \textbf{Edges}  & \textbf{Classes} & \textbf{Features} \\
\hline
Citeseer &3,312 &4,732 &6 &3,703 \\
Cora & 2,708 & 5,429 & 7 & 1,433 \\
Pubmed &19,717 &44,338 &3 &500 \\
\hline
\bottomrule
\end{tabular}
\label{tab_dataset}
\end{table}

We conduct our experiments on three widely used citation networks: Citeseer, Cora, and Pubmed~\cite{sen2008collective}.
Table~\ref{tab_dataset} shows their statistics.
In each dataset, nodes are documents, edges are citations among them, and labels are research topics.
Their features are all bag-of-words features.
Besides RECT-L, we further compare a famous unsupervised method DeepWalk~\cite{perozzi2014deepwalk} and some other supervised methods (LSHM~\cite{jacob2014learning}, RSDNE~\cite{wang2018rsdne}, GCN~\cite{kipf2017semi}, APPNP~\cite{klicpera2019predict} and TEA~\cite{yang2019triplet}).
Following~\cite{wang2018rsdne}, we set the embedding dimension to 200.
For all baselines, we adopt their best hyper-parameters.
In RECT-L and our methods, we all adopt two GCN layers, PReLU activation, mean squared error loss, and Xavier initialization.
We also follow~\cite{yang2015network} to reduce the node feature dimension to 200 via SVD decomposition, and follow~\cite{li2018deeper} to expand the original labeled node set size to $n/\zeta^{\tau}$, where $n$ is the graph node number, $\zeta$ is the average node degree, and $\tau$ is the number of the used GCN layers.
In addition, we fix the training epoch number to 100, adopt Adam SGD optimizer, and use the 200-dimensional outputs of the first hidden layer as embedding results, following~\cite{wang2020RECT}.

\begin{table}[!t]
\small
\caption{Micro-F1 scores on node classification tasks.}
\centering
\begin{tabular}{c||lll|lll|lll}
\hline
 & \multicolumn{3}{c|}{Citeseer} & \multicolumn{3}{c|}{Cora} & \multicolumn{3}{c}{Pubmed} \\
\cline{2-10}
 & 1\% &3\% &5\% &1\% &3\% &5\% &1\% &3\% &5\% \\
\hline
\hline
DeepWalk  &0.1941 &0.2935 &0.3713 &0.1972 &0.3401 &0.4916 &0.3766 &0.5879 &0.6350\\
LSHM &0.1779 &0.2143 &0.2648 &0.1284 &0.1295 &0.2233 &0.3331 &0.3591 &0.3965 \\
RSDNE &0.2291 &0.3066 &0.4035 &0.2465 &0.3869 &0.5167 &0.4193 &0.6219 &0.6862 \\
GCN &0.4194 &0.5211 &0.5478 &0.4756 &0.5984 &0.6266 &0.6067 &0.6479 &0.6664\\
APPNP &0.4192 &0.5397 &0.5692 &0.4921 &0.6380 &0.6791 &0.6036 &0.6287 &0.6514\\
TEA &0.2554 &0.3564 &0.4010 &0.2996 &0.4966 &0.5770 &0.4953 &0.5848 &0.6431\\
RECT-L &0.4506 &0.5754 &0.6204 &0.4964 &0.6564 &0.7325 &0.6679 &0.7495 &0.7668\\
\hline
Ours$_{\mathit{SL}}$ &0.5001 &0.6004 &0.6326 &0.5288 &0.6748 &0.7374 &0.7206 &0.7622 &0.7586\\
Ours$_{\mathit{SUL}}$ &\textbf{0.5343} &0.6228 &0.6497 &0.5125 &0.6761 &0.7275 &0.6641 &0.7419 &0.7336\\
Ours$_{\mathit{SUL}^*}$ &0.5281 &0.6226 &0.6500 &0.4984 &0.6636 &0.7208 &0.6612 &0.7406 &0.7309 \\
Ours$_{\mathit{SL\textit{-}SUL}}$ &0.5297 &\textbf{0.6229} &0.6513 &0.5450 &\textbf{0.6963} &\textbf{0.7515} &0.7224 &0.7704 &0.7688\\
Ours$_{\mathit{SL\textit{-}SUL}^*}$ &0.5293 &0.6226 &\textbf{0.6518} &\textbf{0.5474} &0.6919 &0.7507 &\textbf{0.7353} &\textbf{0.7752} &\textbf{0.7730}\\
\hline
\hline
\end{tabular}
\label{tab_micro-f1}
\end{table}

\subsection{Node Classification}\label{sub_sect_nc}
This experiment follows the same procedure as in~\cite{wang2018rsdne}.
Specifically, we first randomly choose two classes as unseen in Citeseer and Cora, and one class as unseen in Pubmed.
After that, we remove all the nodes from the unseen classes in the training data, and then apply various graph embedding methods.
Finally, an SVM classifier, which is trained based on the resulted embeddings and the original balanced training data, is used to classify the testing nodes.

Table~\ref{tab_micro-f1} reports the classification performance in terms of Micro-F1.
At a glance, we can see the advantage of our label expansion strategies.
Generally, our methods outperform the original RECT-L and other baselines by a large margin in most label settings.
This improvement would become more significant when the training size is very small.
In addition, a very surprising finding is that the performance of Ours$_{\mathit{SUL}}$ is closely related to the performance of Ours$_{\mathit{SUL}^*}$.
This indicates that we can always find discrimination information for unseen classes through clustering, even if the true class number is unknown.
At last, we can find that combing two label expansion results would get the best performance.
This indicates that our two label expansion strategies are complementary for effective embedding learning.

\begin{figure}[t]
\centering
\subfigure{
    \includegraphics[width=0.40\textwidth]{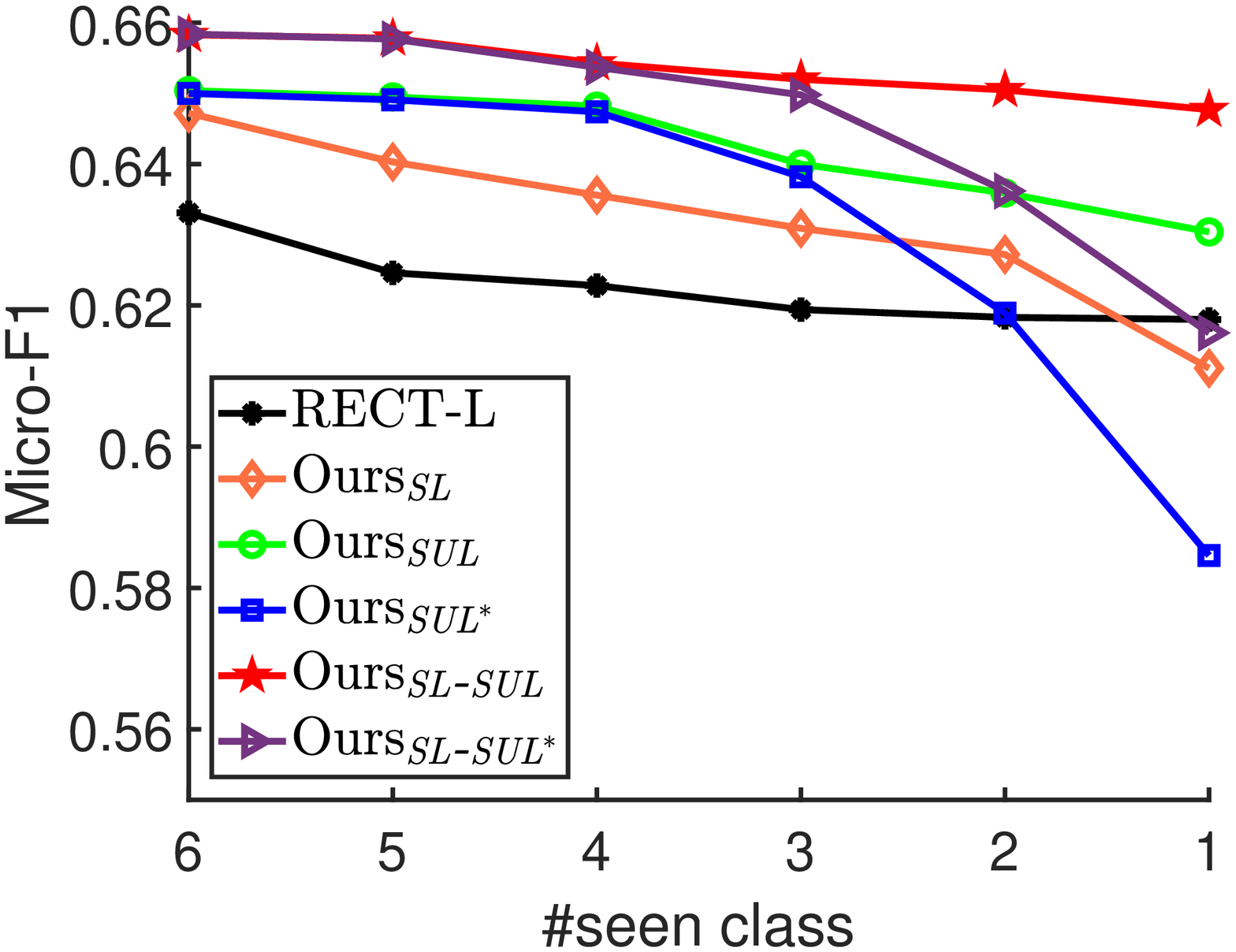}
}
\subfigure{
    \includegraphics[width=0.40\textwidth]{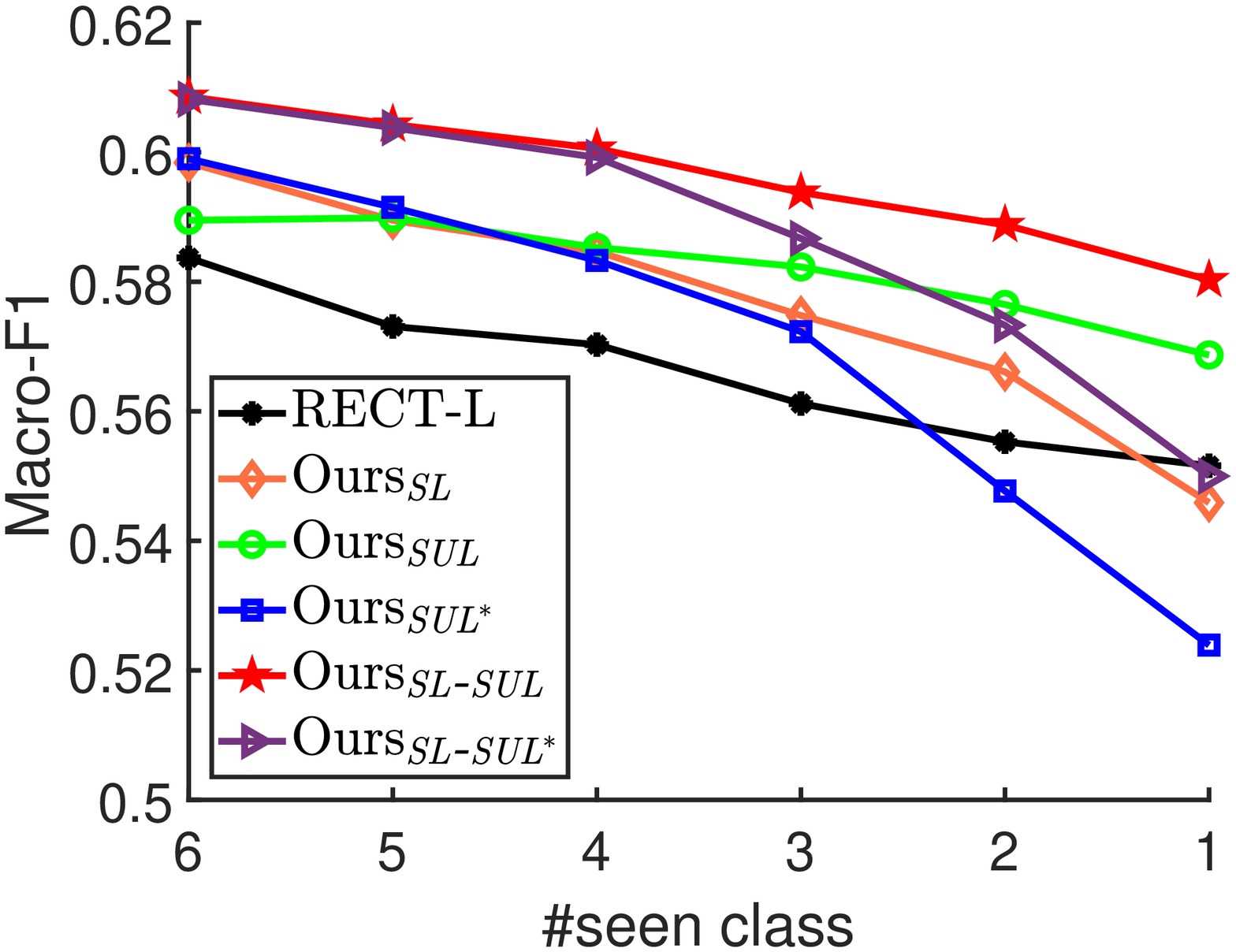}
}
\caption{Classification performance w.r.t. \#seen class on Citeseer with 5\% label rate.}
\label{fig_remove_number}
\end{figure}

\subsection{The Effect of Seen/Unseen Class Number}
We continue to use the Citeseer dataset with 5\% label rate.
As shown in Fig.~\ref{fig_remove_number}, although all the performance declines smoothly when the seen class number decreases, our two label expansion strategies (especially when combing both of them) steadily improve the performance of RECT-L.
This clearly reflects the effectiveness of our methods.

\section{Conclusion} \label{section_conclusion}
In this paper, we give new insights into the mechanisms of RECT and its application in ZGE.
In particular, we analyse its relationship with ZSL methods, and the possible limits that it has.
To fully realize its potentials, we propose two label expansion strategies.
Specifically, we propose to expand the label sets of both seen and unseen classes.
In addition, we also study the theoretical properties of our methods.
Finally, we conduct extensive experiments to demonstrate the effectiveness of our methods.

\subsubsection*{Acknowledgment}
This work is supported in part by National Key D\&R Program of China (2019YFB1600704), National Natural Science Foundation of China (61902020, 61872207), Macao Youth Scholars Program (AM201912), FDCT (FDCT/0045/2019/A1, FDCT/0007/2018/A1), GSTIC (EF005/FST-GZG/2019/GSTIC), University of Macau (MYRG2018-00129-FST), and Baidu Inc.

\bibliographystyle{splncs04}
\bibliography{simple}
\end{document}